%% file: main.tex
\documentclass{acm_proc_article-sp}
\pdfoutput=1
\permission{Copyright \copyright  Simone Cirillo, Stefan Lloyd, Peter Nordin.  http://arxiv.org/licenses/nonexclusive-distrib/1.0/ Abstracting and quoting with credit is permitted. No use for commercial purposes is allowed.}

\usepackage[english]{babel}
\usepackage{placeins}
\usepackage{graphicx}
\usepackage{grffile}
\usepackage{booktabs}
\usepackage{url}
\usepackage[textfont={normalfont, small}, labelfont={bf, small}, skip=15pt, hang]{caption}
\usepackage{mathtools}
\usepackage{amsmath}
\begin{document}

\title{Evolving intraday foreign exchange trading strategies utilizing multiple instruments price series}
\numberofauthors{3}

\author{
\alignauthor
Simone Cirillo\\
       \affaddr{\normalsize\sffamily{Almont Capital LLC}}\\
       \affaddr{\normalsize\sffamily{12800 Hillcrest Road}}\\
       \affaddr{\normalsize\sffamily{Dallas, Texas 75230}}\\
       \email{\normalsize\sffamily{cirillo@almontcapital.com}}
\alignauthor
Stefan Lloyd\\
       \affaddr{\normalsize\sffamily{Almont Capital LLC}}\\
       \affaddr{\normalsize\sffamily{12800 Hillcrest Road}}\\
       \affaddr{\normalsize\sffamily{Dallas, Texas 75230}}\\
       \email{\normalsize\sffamily{lloyd@almontcapital.com}}
\alignauthor			
Peter Nordin\\
      % \affaddr{Chalmers}\\
			 \affaddr{\normalsize\sffamily{Chalmers University of Technology}}\\
       \affaddr{\normalsize\sffamily{Maskingr{\"a}nd 2}}\\
       \affaddr{\normalsize\sffamily{Gothenburg, Sweden}}\\
       \email{\normalsize\sffamily{pnordin@chalmers.se}}
}

\maketitle

\begin{abstract}
We propose a Genetic Programming architecture for the generation of foreign exchange trading strategies. The system's principal features are the evolution of free-form strategies which do not rely on any prior models and the utilization of price series from multiple instruments as input data. This latter feature constitutes an innovation with respect to previous works documented in literature. In this article we utilize Open, High, Low, Close bar data at a 5 minutes frequency for the AUD.USD, EUR.USD, GBP.USD and USD.JPY currency pairs. We will test the implementation analyzing the in-sample and out-of-sample performance of strategies for trading the USD.JPY obtained across multiple algorithm runs. We will also evaluate the differences between strategies selected according to two different criteria: one relies on the fitness obtained on the training set only, the second one makes use of an additional validation dataset. Strategy activity and trade accuracy are remarkably stable between in and out of sample results. From a profitability aspect, the two criteria both result in strategies successful on out-of-sample data but exhibiting different characteristics. The overall best performing out-of-sample strategy achieves a yearly return of 19\%.
\end{abstract} 


\begingroup
\let\clearpage\relax
\include{1_Introduction}
\include{2_GP_System}
\include{3_Setup}

\include{4_Results}
\include{5_Discussion}
\include{6_Conclusions}
\endgroup
\bibliographystyle{abbrv}
\bibliography{references}

\end{document}

%% file: 1_Introduction.tex
\section{Introduction}
\label{sec:intro}

The Foreign Exchange (FX) Market is the most liquid financial market in the world and its average daily turnover is continuously growing. According to the Bank for International Settlements, trading in Forex markets averaged \$5.3 trillion per day in April 2013 \cite{bis2013}. Due to its strong liquidity the FX market is often considered to be the market where it is most difficult to consistently make financial returns from trading.

So far there have been many academic investigations with the aim of developing technical trading strategies for financial markets and products using nonlinear computation techniques such as Artificial Neural Networks \cite{azoff1994neural}, Genetic Algorithms \cite{allen1999using}, and Genetic Programming \cite{potvin2004generating}.
Most often the asset class of choice for developing these models and strategies has been stocks but attempts have been made on the FX market as well.

Generally the obtained results are positive, exhibiting moderate profitability on unseen data. 
However, the applicability and reproducibility of these academic results on real-life implementations running on live markets are often questioned by professional financial practitioners.
In this article we describe an innovative implementation of a Genetic Programming architecture for the evolution of forex trading strategies, with an emphasis on accurate problem modeling to facilitate successive real market application as well as on establishing a reliable methodology to select, from training and validation performance, strategies capable of obtaining profitable results on unseen, new data.

\subsection{Genetic Programming}
\label{subsec:introgp}
Genetic Programming (GP) is a metaheuristic optimization technique belonging to the class of evolutionary algorithms.
Evolutionary Algorithms (EA) take inspiration from the process of biological evolution. At the start of the process a population of candidate solutions to the problem at hand is randomly generated and it subsequently undergoes a process of evaluation, selection, reproduction, crossover and mutation over a number of generations. As generations progress, progressively better solutions are found.

GP has been popularized by John Koza \cite{koza1992genetic} and it has been successfully applied to a variety of problems in many application domains including electronic circuit design, optical lens systems, robotics, game playing, bioinformatics, image and signal processing, scheduling, etc\ldots \cite{koza2010human}.
In GP, the candidate solutions being optimized by means of evolution are programs computing mathematical expressions; the measure utilized to evaluate how good they are at solving the defined problem, driving the process, is called fitness function.
One of the key strengths of GP is that there are no human a priori assumptions made about the model to begin with, as it is the model itself that is being evolved. Therefore the only bias GP is subject to is the inductive bias: the set of assumptions "learned" on the training data.

This contrasts, for instance, with the approach taken by Genetic Algorithms (GA), another type of EA, where the starting point is a model of predetermined structure and only numerical values, usually representing coefficients or weights,  undergo evolutionary optimization.

\subsection{GP, Financial Markets,\\and Trading Strategies}
\label{subsec:gpfx}
Nowadays financial markets are dominated by algorithmic traders, accounting for over 73\% of US equity volumes in 2010 \cite{mckenney2010literature}. Most algorithmic traders rely on relatively few popular models, assumptions and technical indicators for their strategies.

According to Lo's Adaptive Market Hypothesis (AMH) \cite{lo2004adaptive}, in financial markets profit opportunities generally exist however, due to popularization of models, assumptions and indicators used in technical trading, over time these opportunities gradually disappear while others emerge; a behavior also noted by Gencay et al. \cite{gencay2003foreign}. 

In the case of the foreign exchange market, Neely and Weller`s analyses show that profitability of fixed technical rules substantially fluctuates over time \cite{neely2009adaptive, neely2013lessons}, indicating that the foreign exchange market appears to comply, to some extent, to the AMH.
Subscribing to this view, as models and algorithms go in and out of fashion, an arms-race scenario, a metagame, emerges. Strategies are successful if they beat the market itself as well as the market effects caused by actions of other trading strategies, individuating emergent profit opportunities as well as local inefficiencies.

Independently from the AMH, time series of foreign exchange data have been characterized as chaotic, extremely noisy, and non stationary \cite{giles2001noisy}. These considerations lead to the conclusion that, over sufficiently long horizons, a suitable adaptive model, or an adaptive system to construct models should yield better performance than static models or fixed trading rules.

Summarizing, there is a growing body of evidence indicating that the ability to constantly and reliably find different, novel, models and logic for trading strategies is a decisive issue faced by financial practitioners. GP, given its inherent capability of evolving models without human assumptions, intuitively represents a promising methodology for doing so.

The first attempts to apply GP for the evolution of trading strategies for the foreign exchange market date back to the late 1990s. 
In 1997 Neely, Weller and Dittmar attempted to evolve strategies for the interday trading of several currency pairs; they implemented a binary long/short trading model using overnight interest rates as the obtained returns \cite{neely1997technical}. Their results show out-of-sample annualized excess returns from 1\% to 6\%.

A later, more elaborate work from 2003 always by Neely and Weller considers intraday data with a 30 minutes frequency
%and utilizes the continuously compounded excess returns obtained as fitness function
\cite{neely2003intraday}. Their aim here is the investigation of the impact of transaction costs on the final return performance. After performing experiments with varying commission rates as well as without them, they conclude that their GP strategies are able to find predictable patterns in the data but they struggle to produce positive excess returns once transaction costs are factored in.

In 2001, Dempster and Jones developed a GP system evolving indicator-based trading rules \cite{dempster2001real}. The indicators are computed for 15-minute intervals while the strategies trade on a 1 minute frequency, an impressive computational effort for the time. On the period Q1 1994 - Q4 1997 they obtained 7\% annualized returns. Their investigation shows consistent profits for the first three quarters of the out-of-sample period while later performance is much more volatile and prone to losses, providing indirect support to the AMH and indicating that some sort of model re-training is required to guarantee stable performance in the long term.

Hryshko and Downs implemented a hybrid architecture using GA and reinforcement learning to optimize indicator-based entry/exit rules. They obtained a 6\% profit on the EUR.USD over a 3.5 months period \cite{hryshko2006development}.

In 2006 Brabazon and O'Neill proposed the evolution of strategies using a technique called grammatical evolution, a variant of GP based on formal grammars, reporting out-of-sample profits from 0.1\% to 5\% \cite{brabazon2006biologically}.

More recent years have seen the exponential increase in raw computing power of retail hardware, the widespread adoption of architectures capable of parallel computation, the refinement in software tools, and finally the appearance of many retail trading brokers as well as financial data vendors providing intraday or even real-time price quotes. All of these factors laid the foundations for a renewed interest in the subject, making it feasible and practical to perform evaluations of more elaborate models on larger amounts of data.

Hirabayashi et al. utilized GA to optimize the choice of technical indicators for the construction of buy/sell trading rules \cite{hirabayashi2009optimization}. On hourly data for a two year period with rolling window retraining every 3 months they obtain a combined performance for 2005-2008 of 17\% on the USD.JPY, 80\% for the EUR.JPY and 38\% for the AUD.JPY using leverage. However, the corresponding unlevered returns are only of 2\%, -2\%, and 19\% respectively. 

Wilson and Banzhaf proposed an architecture based on Linear Genetic Programming, a variant of GP. Initially they applied it to the stock market on interday \cite{wilson2009prediction} as well as intraday data \cite{wilson2010interday}. Later they focused on interday forex trading \cite{wilson2010interday2}. Their methodology was tested on the CAD.USD, EUR.USD, GBP.USD and JPY.USD on a rolling-window one-year out-of-sample period and they compared the performance obtained by three different fitness functions: one entirely profit based, the other two explicitly attempting to minimize losses as well. Their reported annualized profits range, after accounting for transaction costs, from -9\% to 13\%, depending on currency pair and fitness measure.

Godinho investigated profitability of a GA rule optimizer on currencies whose exchange rate is bounded, as opposed to free-floating \cite{godinho2012can}. He only reports positive out-of-sample returns for the USD.SGD, one of the bounded currencies analyzed. 

Mendes et al. evolved a set of entry and exit technical trading rules using GA \cite{mendes2012forex}. Applied to EUR.USD and GBP.USD data for frequencies ranging from 1 minute to 1 hour, the gross returns on the training set are solidly profitable. However, statistical performance on test, unseen, data is considerably worse and profitability is eroded in the presence of transaction costs.

Loginov proposes a GP variant, called FXGP, based on the co-evolution of a decision tree representing the strategy logic as well as the technical indicators utilized by the tree \cite{loginov2013utility}; FXGP can also employ periodic or trigger condition based retraining to improve profit consistency on out-of-sample data. Such retraining appears to be fundamental to the approach's profitability: algorithm runs without any sort of retraining are shown to have at most a 38\% chance of any profitability at the end of a 3-years period.

A different way of framing the problem is not the direct evolution of trading strategies, but rather the evolution of models for price prediction; the predictive model is then coupled with a fixed logical rule to produce trading signals.
Evans et al. developed a price forecasting model utilizing GA to optimize the structure of an Artificial Neural Network (ANN) \cite{evans2013utilizing}; they report annualized returns of 23.3\% without the inclusion of transaction costs and with an out-of-sample dataset length of two months.

Vasilakis et al. developed GP models for the 1-day-ahead forecasting of the daily ECB fixing of the EUR.USD rate \cite{vasilakis2012genetic}. The trading performance is shown to outperform a buy-and-hold benchmark strategy, Moving Average Convergence Divergence (MACD) trading, and Evolutionary Artificial Neural Network (EANN) models, a technique combining GA and ANN.

Manahov and Hudson also attempted to forecast intraday forex rates using GP \cite{manahov2013new}. The peculiarity of their approach is that they rely on an entirely simulated marked where the GP individuals compete against each other. In their experiments they utilize 5-minutes data and they show out-of-sample profitability outperforming conventional autoregressive models for the EUR.USD, USD.JPY, GBP.USD, AUD.USD, USD.CHF, and USD.CAD pairs with returns ranging from 5.90\% to 3.76\%. In a more recent study using 1 minute data, their results don't show significant improvement \cite{manahov2014does}.

\subsection{Aim and Scope}
\label{subsec:aimscope}

In this paper we will describe an architecture to evolve strategies for intraday trading in the FX market using Genetic Programming, with the final goal of employing them for live trading.
Our approach features innovative key aspects introduced with the aim of boosting the GP strategies' performance, increase their flexibility, as well as narrowing the gap between the evolutionary optimization process, occurring within the GP environment, and subsequent usage in online, live trading.

For instance, in contrast to documented work, our implementation does not rely on technical indicators, autoregressive inputs, or a pre-defined set of rules but instead it evolves completely free-form trading strategies considering price\\quotes from different forex pairs as input values. Our architecture therefore operates on a significantly larger search space for both problem and solution domains.

Finally, strategies are evaluated within a 3rd party commercial trading software to verify the correctness of the obtained results and to perform further testing.

We will present results obtained by strategies produced by our system for trading the USD.JPY spot rate, with USD as the base currency for calculating performance; we will benchmark them against a buy-and-hold strategy as well as the Barclay's BTOP FX index \cite{btop2014}.
With the aim of establishing a reliable methodology, we will also compare two different criteria for the selection of strategies likely to perform well on unseen data. Good a priori chances of successful generalization is an issue of crucial importance for making feasible the consistent utilization of GP strategies in a live, production, trading setting.

In Section 2 we describe the implementation of the forex-trading Genetic Programming architecture utilized for the evaluations presented in this article. Section 3 outlines the details and the scope of the GP algorithm runs we performed. In Section 4 we present the obtained results which will be discussed and compared against previous work in Section 5. In Section 6 we will be making concluding remarks as well as indicating possible future steps for this line of research.

%% file: 2_GP_System.tex
\section{Forex Trading GP System}
\label{sec:system}

A Genetic Programming architecture is a complex system encompassing many abstraction layers: from the proper GP algorithm implementation, to what the evolved programs are and how do they operate, to finally what the fitness function evaluating their performance is.

\subsection{GP Framework}
\label{subsec:sysoverview}
The core of our system is the HeuristicLab platform, an open-source, extensible plugin-based optimization framework for heuristic and evolutionary algorithms developed and maintained by the Heuristic and Evolutionary Algorithms Laboratory of Upper Austria University of Applied Sciences \cite{wagner2014architecture}.

HeuristicLab enforces by design the separation of logical abstractions and source code modules for the optimization algorithm from those concerning the problem to be solved, facilitating reuse of existing functionalities. Other advantages include access to the entire original codebase, allowing for the reimplementation or addition of any feature; the support for parallel execution, greatly reducing the computation times for the algorithm runs; and finally the powerful built-in solution and population analysis capabilities.

HeuristicLab does support Genetic Programming on its own, specifically in the form of Tree GP: an implementation popularized by Koza \cite{koza1992genetic}. Additional modules for significantly improving the algorithm run times on large dataset, large population setups \cite{Cirillo2014}, as well as functionalities for simulating the financial market and evaluating trading solutions were instead developed by the authors.

%\newpage
\subsection{GP Strategies}
\label{subsec:gpstrats}

The AMH seems to indicate that established technical trading methods, as their popularity grows, tend to lose effectiveness. Our investigation therefore attempts to find profitable trading strategies departing from conventionally used models.

Many of the articles mentioned in Section \ref{subsec:gpfx} employ the techniques of GA and ANN. Such techniques optimize pre-determined models or have very limited model structure optimization capabilities. The works of Neely and Weller \cite{neely1997technical, neely2003intraday},  Wilson and Banzhaf \cite{wilson2010interday2}, Vasilakis et al. \cite{vasilakis2012genetic}, Loginov \cite{loginov2013utility} and Manahov \cite{manahov2013new, manahov2014does}, implementing proper GP, instead allow the optimization process to freely evolve the model structure as well.

An aspect present in every documented previous implementation is the reliance, or the usage, on common technical analysis indicators such as MACD, Momentum, etc\ldots These indicators have their usefulness if interpreted and analyzed by a human trader. However, it is debatable whether an automatically constructed trading strategy would, optimally, evolve towards using them as well since they represent a man-made, inherently biased, perspective. Additionally, given the widespread usage of such technical indicators, their availability to the GP programs might provide a path of least resistance to local optima in the fitness landscape representing conventional, but potentially suboptimal, trading strategies.

Concluding, our architecture will evolve GP free-form strategies that do not make use of conventional technical indicators.

The function set for the experiments presented in this article therefore only consists of arithmetic and trigonometric operators, as well as boolean operators and flow control statements to allow for the emergence of functioning if-then-else constructs. These latter enable the evolved individuals to develop different behaviors in response to changing conditions in the underlying market.

Specifically the members of the function set are: \textit{Addition}, \textit{Subtraction}, \textit{Multiplication}, \textit{Division}, \textit{Sine}, \textit{Cosine}, \textit{Tangent}, \textit{IfThenElse}, \textit{GreaterThan}, \textit{LessThan}.
The terminal set instead consists of the input variables, \textit{Variable}, as well as randomly-generated constants, \textit{Constant}; both types of terminal node allow for a multiplicative weight value.

Trees have both a depth limit and a length limit, so to reduce the phenomenon of bloat \cite{altenberg1994emergent, banzhaf1998geneticch7}. 
The trees representing our GP individuals are generated using a probabilistic method.
Crossover is implemented using subtree-swapping. Selection is tournament-based.
Mutation can occur in the following ways: change of the type of non-terminal node, change of the value of the weight of a terminal node, removal of a sub-tree, replacement of a sub-tree with a newly generated one.

\vfill
\pagebreak
\subsection{Strategy Workflow}
\label{subsec:workflow}

\subsubsection{Input Data}
\label{subsubsec:inputdata}
%In many cases the results are based on end-of-day price observations \cite{neely1997technical, brabazon2006biologically, evans2013utilizing, vasilakis2012genetic}.
In fast moving markets exhibiting high volatility, interday strategies obtain very limited informational exposure and trading opportunities. Their potential profitability results therefore at a disadvantage in comparison to the growing number of faster market actors. For this reason our work will be focusing on generating strategies having a successful and active trading profile on intraday price variations. The risk this choice entails consists in the increased noise intraday data exhibit compared to interday series; meaningful patterns are harder to find and learn.

The data we start from consists of collected tick-level spot pricing data from Citigroup; the tick data is aggregated in the form of 5-minute bars, each bar consisting of four values: \textit{Open}, \textit{High}, \textit{Low}, and \textit{Close} prices for the time interval it represents.

All the works mentioned in Section \ref{subsec:gpfx} only utilize price quotes referring to the same instrument being traded as input data.
An additional innovation of our proposed approach consists in that, while the strategies presented in this work will only trade the USD.JPY, they will be given input bar data for the following instruments: EUR.USD, USD.JPY, GBP.USD, AUD.USD, accounting for a significant part of the total global daily liquidity.

The goal is enabling agents to find inter-security patterns and correlations which may result beneficial to their trading performance. In comparison to having single security price or bar values this substantially increases the dimensionality of the problem space: strategies have a total of 16 input values per datapoint as opposed to the 4 they would have if given inputs only for the traded instrument. However the implicit, emergent, feature selection property of GP means that each strategy only utilizes an arbitrary subset of the total available inputs \cite{banzhaf1998geneticch5}.

Considering the high dimensionality of the problem space and the many-to-many non-trivial correlations and relationships between the variables, we opted not to perform any pre-processing or normalization procedure on the data; bar price values are used as-is. The evolved strategies are therefore exposed to the non-stationarity of the underlying market: strategies might lose predictive or action capabilities if the input values, over time, diverge too much from the ranges they had in training dataset. On the other hand a no-preprocessing policy prevents any loss of information present in the data, leaving much more freedom and therefore pattern learning potential to the evolutionary process. 

\subsubsection{Strategies' Output}
\label{subsubsec:output}

The most common implementation involves mapping the GP program output to a \textit{buy}, \textit{sell}, or \textit{stay} action or to a binary \textit{long}/\textit{short} position; alternatively, the evolved programs only produce a next-step price prediction.
Both of these output types are then coupled to a fixed model, not subject to evolutionary optimization, aware of equity amount and current position responsible for issuing trading orders.
The main limitation of both of these paradigms consists in the fact that the strategy's final performance is entirely dependant by the trading model in place.

In order to provide maximum freedom to the evolved strategies, we will be adopting an implementation similar to the one presented in \cite{manahov2013new, manahov2014does}, where for each set of input values the GP individual computes an output value interpreted as the desired position exposure. 
Valid exposures range from +100 to -100, from a full long position to a full short position.

The main advantage this choice confers is the total decoupling of the strategies from the amount of equity capital traded, as the output value represents an entirely relative quantity. Another emerging characteristic is the evolution of strategies whose output can be seen as a datapoint-by-datapoint confidence value signal, as opposed to a trigger of an imminent trend shift. Both traits are desirable for application in live trading, making it possible to later use the strategy logic with different trading or portfolio management models. 

\subsubsection{Trading Model}
\label{subsubsec:tradingmodel}
In foreign exchange trading, as opposed to the stock market for instance, a full long position means that all the assets are in the quote currency while a full short position conversely means they are all in the base currency. Since one currency goes up the other in the pair goes down, forex trading involves appropriately moving assets from a currency to the other and viceversa. To capture this market dynamic therefore our strategies are evolved in an environment where the only safe position is 50\% long and 50\% short: where the gains made one side are exactly matched by the losses on the other side. 

This modelization successfully makes strategies able to trade on both sides of the market. It also prompts them to action: the total value of a starting 50-50 position will not change no matter the underlying market conditions. A 100-0 or 0-100 starting state would instead be equivalent to a buy-and-hold strategy and could therefore result in profits at the end of the evaluation period. This leads to rewarding of undesired strategy behavior; preliminary experiments confirmed the implicit evolutionary advantage this confers to individuals exploiting the aspect, making it harder for the algorithm to reliably find strategies with an active trading profile.

Strategies are initialized with an arbitrary equity amount placed in a 50-50 position. As explained in Section \ref{subsubsec:output}, once per datapoint the strategy computes its desired position exposure.
If the strategy output differs from the position currently held by more than 10\% a corresponding buy or sell order is issued; in accordance to Citigroup's live forex trading platform, order sizes are always rounded to multiples of 5,000 currency units. Issued orders are executed as market orders on the current close price for the traded security. This makes the strategies' internal logic the only responsible for the performed trading actions.

As the results of Neely and Weller \cite{neely2003intraday}, Mendes et al. \cite{mendes2012forex}, and Godinho \cite{godinho2012can} show, transaction costs have a tremendous impact on strategy profitability.
However they are often ignored or their application is approximate; furthermore no previous article mentions whether their application occurs a priori, so that the evolutionary process implicitly adapts the strategies to their presence, or a posteriori, necessarily resulting in strategies that will perform worse when their activity is accounted for commissions no matter their magnitude.

In this work transaction costs matching those applied by Citigroup on live trading are applied. In our case the transaction cost amount to 15 USD per million dollars transacted, appropriately converted for non-USD based currency pairs.

Finally, execution details are stored to be able to compute various statistics at the end of the evaluation. Our trading simulation is also aware of the datapoint timestamps, allowing it to track trading days and compute performance statistics on a daily basis.
	
\subsubsection{Fitness Evaluation}
\label{subsubsec:fitness}

Traditionally, GP is used on problems that from a computational perspective have been framed in terms of regression or classification problems: in such cases the fitness function is intuitively and aptly defined as an error metric, e.g. Mean Square Error, for the desired output values known beforehand. However, in simulation-based problems such as financial trading it is pointless to define a desired output value for every single datapoint: the consequence of an action performed at time $t$ (entering a position), will not be known until at least time $t+1$ (exiting the position). Therefore a proper fitness measure cannot be defined as an error on a per-datapoint basis, but instead as a measure of performance obtained over the entire dataset.

A natural performance measure for trading strategies is profitability: how much profit the strategy makes over time. Another objective that some investigations attempt to promote via explicitly factoring it in the fitness function is the minimization of incurred losses.

Wilson and Banzhaf in \cite{wilson2010interday2} compare a raw profits fitness with two others factoring in the maximum drawdown (MDD, the maximum cumulative loss since the start of the period): one subtracts the MDD from the final value of assets held, the other, more conservative, instead divides this final value by the MDD. In most of their evaluations the most conservative fitness generated the highest profits.
Loginov utilizes a fitness in pips, the minimal profit unit in forex trading, multiplied by a quantity dependant on the number of stay, buy, and sell actions to obtain desired trading activity levels \cite{loginov2013utility}.
Hryshko and Downs \cite{hryshko2004system}, Mendes et Al \cite{mendes2012forex}, and Dempster and Jones \cite{dempster2001real} all utilize the Stirling ratio: the profit divided by the MDD, or a variation of thereof.

We do agree on considering low drawdowns an important characteristic for the application on live markets; however, given the many innovative aspects, our main concern with this work is evaluating the viability of our approach from the perspective of pure profitability.
Taking into account the relative nature of the trading signals emitted by our strategies, not tied to any specific equity amount, we can define:\\
\begin{equation}
\label{eq:returns}
\mathit{return} = \frac{\mathit{Final\;NAV}}{\mathit{Initial\;NAV}} - 1
\end{equation}

Where $\mathit{NAV}$ is the Net Assets Value of the strategy, the total value of assets held.
This expression, for values of $\mathit{Initial\;NAV} > 0,$ would be suitable for use as a fitness function for GP individuals. However an earlier, more limited version of our systems had very strict constraints on the fitness function such as not allowing negative values as well as applying a minimization policy to the scores, so lower scores were considered better. Therefore we had to formulate a function complying to those criteria, we later decided to keep the function in use to be able to compare performance of the old GP engine with the current one.
The fitness function used in this work to evaluate strategies is therefore defined as:
\begin{equation}
\label{eq:fitness}
\mathit{fitness} = e^{ - \mathit{return}}
\end{equation}

This function maps to a monotonically decreasing fitness landscape with no singular or negative values, a clear limit value for the worst case scenario where the strategy lost all of its money:
%\begin{equation*}
	$$\lim_{\mathit{Final\;NAV} \to 0} \mathit{fitness}=e$$
%\end{equation*}

as well as clearly showing whether the strategy is profitable ($\mathit{fitness}<1$) or not ($\mathit{fitness}>1$).
In order to promote strategies with an active trading profile rather than just taking advantage of long term trends in the market we introduced a number of minimum trades, defined as position entry-exit pairs, the strategy has to perform.  If the strategy trades fewer times than the specified amount it is assigned a very penalizing fitness score. Another case resulting in penalization is for strategies that at the end of their evaluation on the training dataset lost all, or more, of the equity they started with.

%% file: 3_Setup.tex
\section{Experimental Setup}
\label{sec:setup}

\subsection{Population Size and Generation Count}
\label{subsec:popgen}
As described in Section \ref{subsubsec:inputdata}, the problem space our strategies evolve on has considerably larger dimensionality than previously documented attempts and the underlying data series are non-stationary. In addition, financial trading is a simulation problem (as opposed to regression or classification problems), therefore not allowing for per-datapoint optimal output values univocally defining a priori the desired behavior. Finally, our main concern is finding individuals that would exhibit good performance on unseen data so overtraining is a potential issue.
 
Such conditions imply that an exhaustive sampling, and learning, of the search space would be unfeasible. For these reasons we will then greatly favor exploration of the search space over exploitation, a conclusion which Dempster and Jones \cite{dempster2001real} as well as Loginov \cite{loginov2013utility} also explicitly reach. Therefore we will adopt a very high population, low generation setup.

In contrast, previous investigations on the subject favored low population and high generation counts instead, favoring exploitation. To exemplify, early works such as \cite{neely1997technical} and \cite{brabazon2006biologically} only have a population size of 500 and a generation count of 50; an understandable choice considering the computational resources available at the time. The largest documented population setting so far instead is 10,000 in \cite{manahov2014does}.

In this work we utilize a population size of 75,000 and a generation count of only 15.

\subsection{Datasets}
\label{subsec:datasets}
The algorithm runs and performance evaluations featured in this work make use of three different datasets: Training, Validation, and Out-of-Sample. As mentioned in Section \ref{subsubsec:inputdata} the datasets are constituted by 5-minute bars for the EUR.USD, USD.JPY, GBP.USD, and AUD.USD, from Sunday at 17:00 to Friday at 17:00 Eastern Time.

The training set constitutes the dataset all individuals are evaluated on and the fitness score, as per Equation \ref{eq:fitness}, obtained on it is the only one guiding the evolutionary algorithm.
On preliminary investigations we found that utilizing a short time period for training increases the emergence of overtrained strategies, failing to learn generalized patterns and behaviors. Such strategies typically focus on specific or unique profitable short-term trends in the training market but exhibit very little trading activity or, worse, disastrous performance when run on different data. To counter this phenomenon the training set we consider consists of ten months of data, for a total of 87,603 simulation datapoints.

When the evaluation of strategies on the validation set is required, it is performed only for the top-scoring 10\% of the population. Given the large population sizes we employ, this dramatically reduces execution times for algorithm runs while not negatively affecting the chance of finding enough profitable, successful strategies on the validation set as well. The validation set consists of the two months worth of data immediately successive to the training set, 17,612 datapoints.

To test the generalization capabilities of the evolved agents, after the breeding process we evaluate the best strategies on a third dataset encompassing a time period successive to their training and validation sets. This Out-of-Sample (OoS) set consists of one year of 5-minute bar data, or 105,759 simulation datapoints, directly following the time periods considered for training and validation.

Table \ref{tab:datasetsummary} resumes the datasets' details.

\begin{table}
	\centering
	\tiny
	\begin{tabular}{llrr}
	\toprule
		Dataset&Period&Days&Datapoints\\
	\midrule
		Training&23 February 2012 - 23 December 2012&213&87,604\\
		Validation&24 December 2012 - 22 February 2013&40&17,612\\
		OoS&23 February 2013 - 25 February 2014&254&105,759\\
	\bottomrule
	\end{tabular}
	\caption{Dataset Details}
	\label{tab:datasetsummary}
\end{table}

%Somewhere, add considerations on oos set size (probably in results' discussion)

A commonly used policy is rolling-window retraining \cite{dempster2001real, wilson2010interday2, loginov2013utility}: agents are initially evolved on the training set, then applied to a test set. Successively additional algorithm runs are performed by progressively shifting forward, usually partially overlapping with the preceding run, the training and test sets. Finally, the partial out-of-sample results from each train-test cycle are combined to obtain final performance. 

Since financial time series are not stationary processes \cite{giles2001noisy}, this approach has been shown to boost profitability \cite{loginov2013utility}. However the focus of this work is the investigation of the long-term predictive capabilities of the proposed architecture and of its statistical reliability at finding profitable strategies; therefore we opted not to employ rolling-window retraining in the presented evaluations.

%Identifying systematic criteria for evolving trading strategies exhibiting good trading activity levels and return performance beyond the training and the eventual validation datasets.  many investigations adopt a rolling window strategy with periodic or condition-based retraining. While this approach guarantees ongoing re-adaptation of the system, alone it fails to give insight on the statistical reliability of the training setup and algorithm run settings at producing strategies whose performance will be acceptable for live use.
\subsection{Strategy Selection Criteria}
\label{subsec:stratselection}
In order to select strategies to evaluate on the Out-of-Sample set we propose two alternative selection criteria. The first is simply the fitness score, defined in Equation \ref{eq:fitness}, obtained on the training set.

The second criterion is a combined value calculated from the fitness scores on both the training and validation sets if these are both profitable, with values $<1$.
The formula for the combined score is:
\begin{equation}
\label{eq:combiscore}
s(f_{t}, f_{v}) =  |f_{t} - f_{v}| + 1 - \frac{\sqrt{(1-f_{t})^2 + (1-f_{v})^2}}{\sqrt{2}}
\end{equation}

\begin{figure}
	%\centering
	\includegraphics[width=0.5\textwidth]{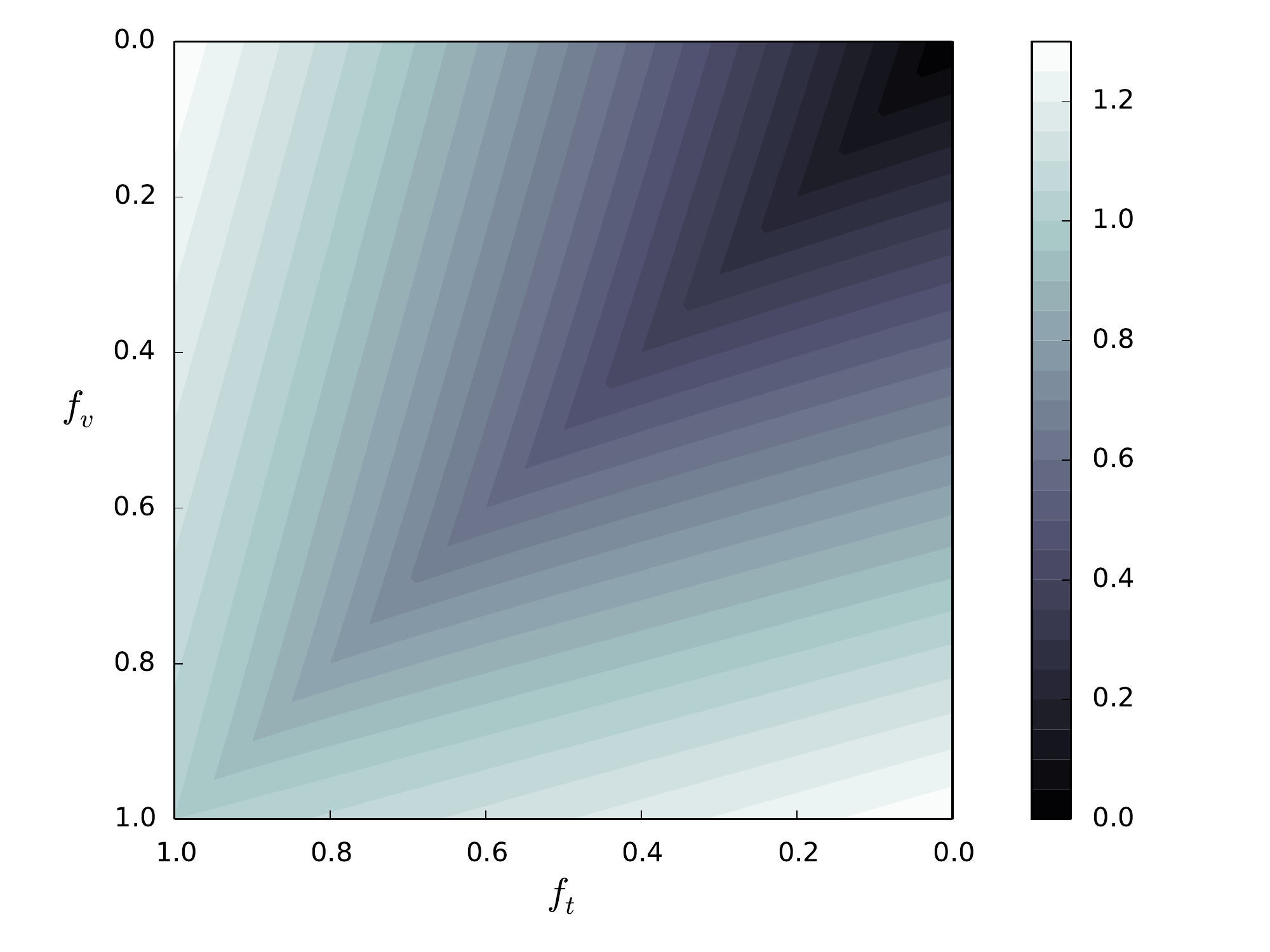}
	\caption{Contour plot for the combined training and validation scoring function listed in Equation 3}
	\label{fig:combi}
\end{figure}

$f_{t}$ and $f_{v}$ are the training and validation fitness scores, respectively. 
It is the sum of the function
\begin{equation*}
d(f_{t}, f_{v}) = |f_{t} - f_{v}|
\end{equation*}

the absolute difference of the two fitness scores, with the function
\begin{equation*}
r(f_{t}, f_{v}) = 1 - \frac{\sqrt{(1-f_{t})^2 + (1-f_{v})^2}}{\sqrt{2}}
\end{equation*}

the circular, infinite cone with apex $(1,1)$, extending into the positive half-space with respect to the Z axis, and scaled to have roots on the circle passing for the origin of the $f_{t}, f_{v}$ plane.

The function, illustrated in Figure 1, provides a landscape with a valley that favors individuals with low difference between their scores on the two sets; unless this difference is already null, however, it is still possible to obtain better combined scoring values if the difference increases due to a significant improvement of only one out of the two sets' score.
To keep consistency with training-only scoring, this combined formula was also designed to minimize scores of better performing strategies.
Given the equal weighting between the $f_{t}$ and $f_{v}$, this combined scoring is particularly suited to instances where the training and validation set have equal length.  

It is worth clarifying that this combined measure does not replace the training-only score for GP fitness purposes, but it is only used for selecting the strategies to run on the Out-of-Sample set. In other words neither the fitness score obtained on the validation set nor the combined score have any impact on the evolutionary process. It is always only the performance on the training set that drives the GP optimization. 

\subsection{Out-of-Sample Evaluation and Analytics}
\label{subsec:oosevaluation}
The Out-of-Sample evaluation is not performed internally to the HeuristicLab GP environment but within a commercial software for quantitative trading where an execution flow analogous to the one occurring within the breeding environment is implemented.
Custom-written modules for HeuristicLab are capable of parsing and saving the trees representing GP individuals into a linearized, plain-text, C\# syntax form which the 3rd party trading software is capable of importing and running.

It is also possible to re-run strategies on their original training and validation datasets. The comparison of the two execution flows in the different environments allows for an external, independant, verification of the correctness of the trading, order execution, NAV, and position tracking models used in the breeding process.

Additionally, utilizing the quantitative trading external software makes it possible to obtain a vast number of performance statistics for the strategies, allowing a more thorough analysis. If computed within the HeuristicLab environment these additional statistics would noticeably increase the already intensive algorithm running times.
Finally, the external software is capable of connecting to real-time market data providers and trading brokers. This means that, if desired, it is already technically and practically possible to run the strategies evolved by our system on live, actual, markets.

%% file: 4_Results.tex
\section{Results}
\label{sec:results}
\FloatBarrier
A total of five GP algorithm runs was performed; for each run and selection criterion we considered the top performing 10 individuals. Table \ref{tab:runsettings} summarizes the GP and trading problem setup we used to evolve the strategies. We will now present the results obtained on the Training and on the Out-of-Sample (OoS) dataset for the trading strategies selected by the two criteria as well as analyze the length and structure of their expression trees.  

\begin{table}
	\centering
	\begin{tabular}{lr}
	\toprule
	Population&75,000\\
	Generations&15\\
	Crossover Rate&90\%\\
	Mutation Rate&15\%\\
	Max Tree Depth&8\\
	Max Tree Length&60\\
	Elitism&1\\
	Fitness Function&$\mathit{fitness} = e^{-\mathit{return}}$\\
	Minimum Trades&50\\
	Instrument Basket&AUD.USD, EUR.USD\\
									 &GBP.USD, USD.JPY\\
	Traded Instrument&USD.JPY\\
	Validated Agents&Top 10\%\\
	Selected Agents&Top 10\\
	\bottomrule
	\end{tabular}
	\caption{Common GP Run Settings}
	\label{tab:runsettings}
\end{table}

\subsection{Training Dataset Performance}
\label{subsec:trainingresults}
\begin{figure}
\setlength{\abovecaptionskip}{0pt}
\centering
	\includegraphics[width=0.5\textwidth]{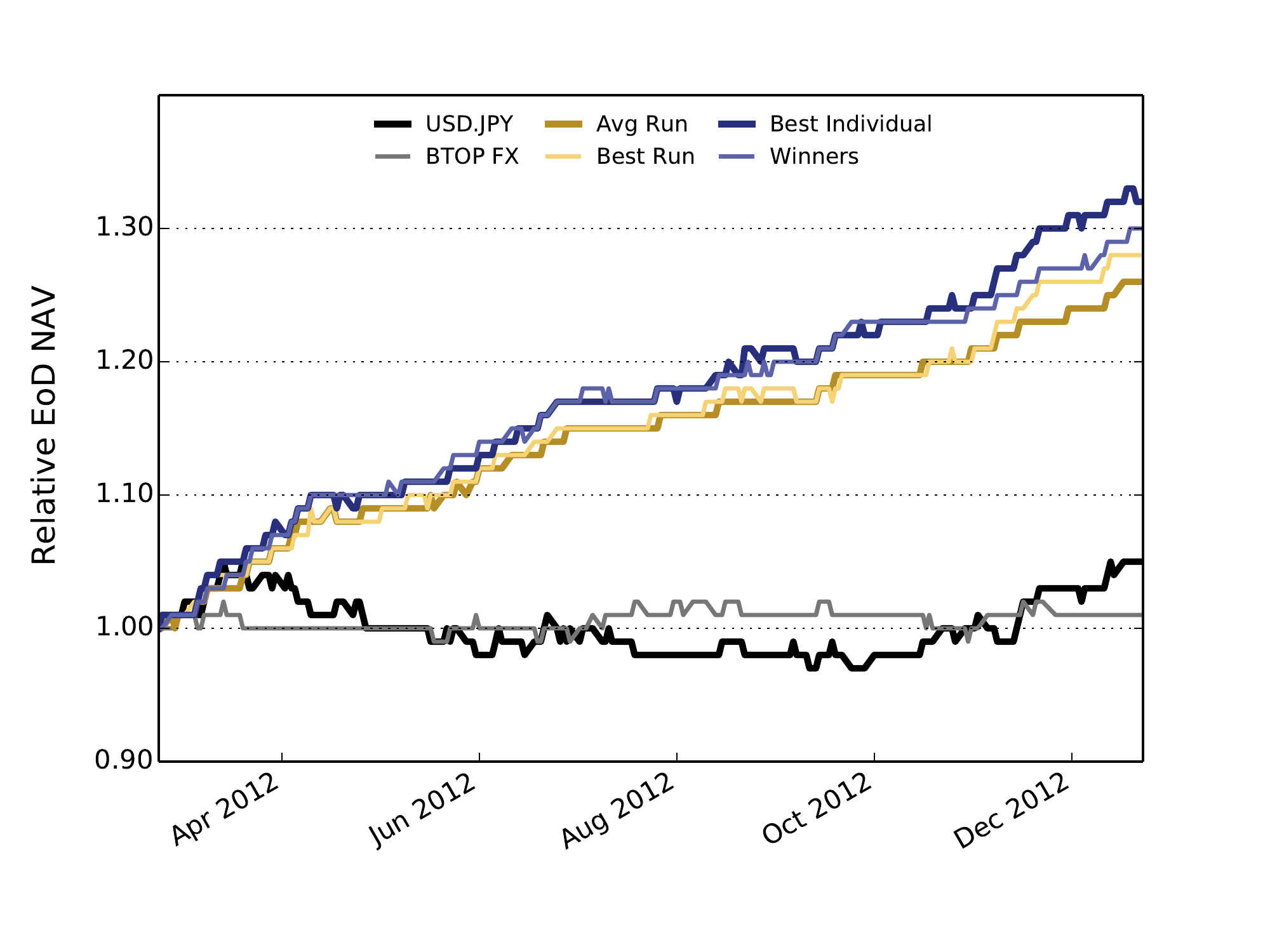}
	\caption{Training set End-of-Day NAV, \textit{Tr} criterion strategies}
	\label{fig:tset_trainonly}
\end{figure}

\begin{figure}
\setlength{\abovecaptionskip}{0pt}
\centering
	\includegraphics[width=0.5\textwidth]{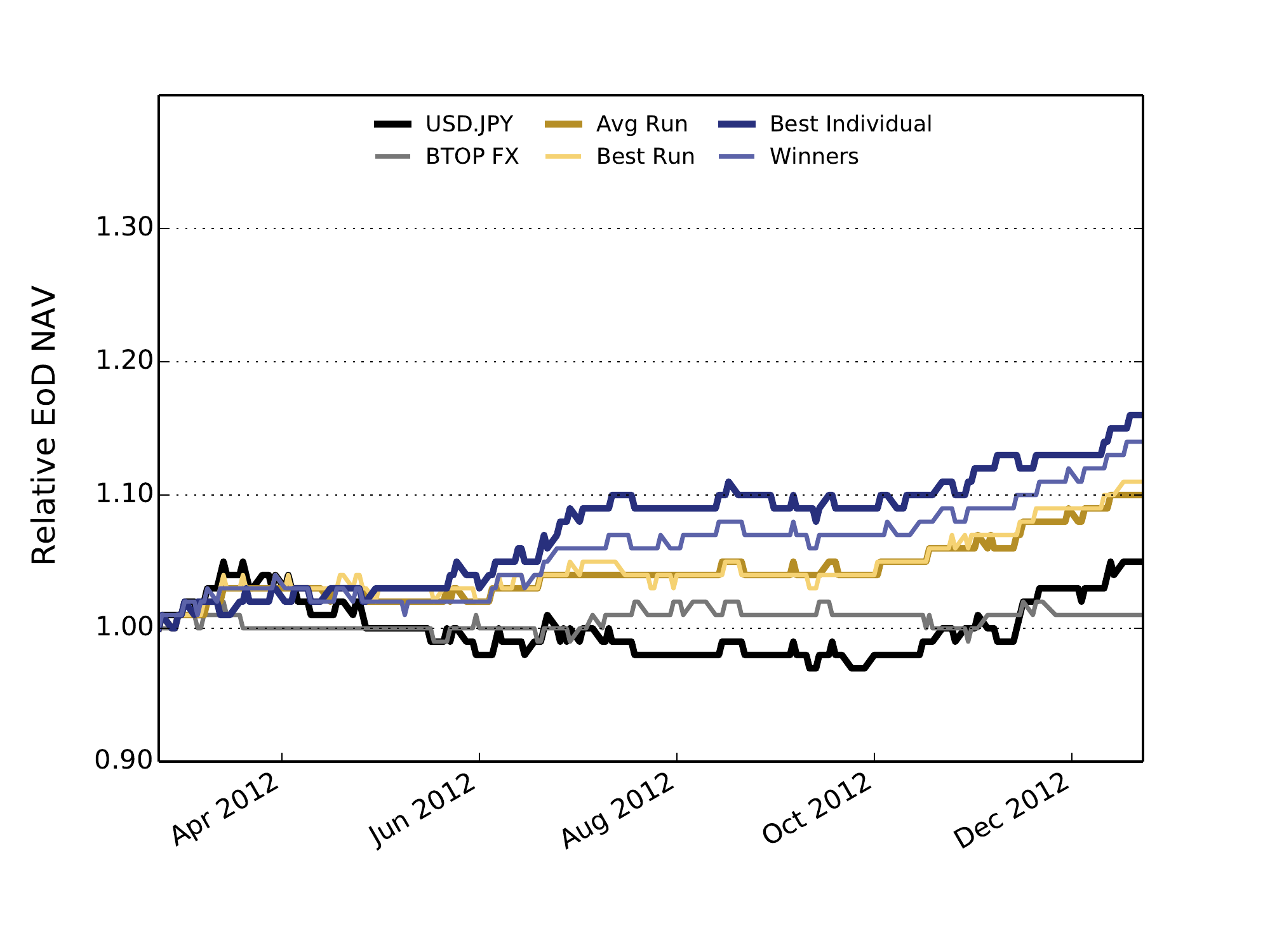}
	\caption{Training set End-of-Day NAV, \textit{TrVa} criterion strategies}
	\label{fig:tset_traintest}
\end{figure}

Figure \ref{fig:tset_trainonly} shows the End-of-Day (EoD) relative NAV values, expressed as $$\frac{\mathit{Current NAV}}{\mathit{Initial NAV}} * 100$$ on the training dataset for the training only selection criterion (\textit{Tr}).
The graph shows the single best performing strategy obtained across the 5 runs (Best Individual), the average of the single best performers of the 5 runs (Winners), the best average of the 10 strategies considered in a run (Best Run), and the average of all the 50 strategies obtained across the 5 runs (Avg Run). For benchmarking purposes the graph includes the USD.JPY price curve, analogous to a buy-and-hold strategy as well as the Barclay's BTOP FX index.
Figure \ref{fig:tset_traintest} instead shows the EoD NAV values obtained by the strategies individuated by the combined selection (\textit{TrVa}) according to the same groupings as Figure \ref{fig:tset_trainonly}.

Tables \ref{tab:tset_trainonly_daily} and \ref{tab:tset_traintest_daily} display the final return (Return), the ratio of profitable days (Days $>0$) and Pearson correlation coefficient ($\rho$) with the daily returns of the USD.JPY. The values are reported for the two financial benchmarks (USD.JPY, BTOP FX), the average of all the 10 considered agents for each run (Run 1-5), the single best performing agent (Best Agent), the average of the single top performers for the five runs (Winners), and the total 50 strategies average (Avg Run). Table \ref{tab:tset_trainonly_daily} refers to the \textit{Tr} criterion while Table \ref{tab:tset_traintest_daily} to \textit{TrVa}.

Table \ref{tab:tset_activity} reports, always for the strategies' Training dataset, various aggregate statistics about the performance and trading activity of the 50 strategies for the two criteria. The Performance section shows the number of profitable run averages and agents as well as those beating a buy-and-hold trading strategy and the global average daily return. The Trading Activity section displays the average number of trades (Trades), winning trades (Winning) and long side trade (Long) ratios as well as the highest achieved winning and long trades ratios (Max Winning, Max Long).

\begin{table}
	\centering
	\footnotesize
	\begin{tabular}{lrrr}
	\toprule
		&Return&Days $>0$&$\rho$\\
	\midrule
		\textbf{Benchmarks}&&&\\
		&&&\\
	%\midrule
		USD.JPY&1.05&51.20\%&1.00\\
		BTOP FX&1.01&51.50\%&0.30\\
	\midrule
		\textbf{Runs}&&&\\
		&&&\\
	%\midrule
		Run 1&1.27&68.08\%&0.19\\
		Run 2&1.26&68.54\%&0.05\\
		Run 3&1.26&67.61\%&-0.07\\
		Run 4&1.26&66.20\%&-0.05\\
		Run 5&1.23&60.56\%&-0.07\\
	\midrule
		%\textbf{Aggregates}&&&&&\\
	%\midrule
		Best Agent&1.32&66.67\%&0.25\\
		Winners&1.30&72.30\%&-0.10\\
		Avg Run&1.26&75.12\%&0.02\\
		\bottomrule
	\end{tabular}
	\caption{Daily returns statistics on the Training set, \textit{Tr} strategies}
	\label{tab:tset_trainonly_daily}
\end{table}

\begin{table}[h]
	\centering
	\footnotesize
	\begin{tabular}{lrrr}
	\toprule
		&Return&Days $>0$&$\rho$\\
	\midrule
		\textbf{Benchmarks}&&&\\
	%\midrule
	&&&\\
		USD.JPY&1.05&51.20\%&1.00\\
		BTOP FX&1.01&51.50\%&0.30\\
	\midrule
		\textbf{Runs}&&&\\
	%\midrule
	&&&\\
		Run 1&1.10&60.09\%&0.82\\
		Run 2&1.11&53.99\%&0.94\\
		Run 3&1.10&51.64\%&0.95\\
		Run 4&1.11&53.99\%&0.90\\
		Run 5&1.08&50.70\%&0.95\\
	\midrule
		%\textbf{Aggregates}&&&&&\\
	%\midrule
		Best Agent&1.16&56.81\%&0.40\\
		Winners&1.14&58.69\%&0.78\\
		Avg Run&1.10&51.17\%&0.96\\
		\bottomrule
	\end{tabular}
	\caption{Daily returns statistics on the Training set, \textit{TrVa} strategies}
	\label{tab:tset_traintest_daily}
\end{table}

\begin{table}
	\centering
	\begin{tabular}{lll}
	\toprule
	&\textit{Tr}&\textit{TrVa}\\
	\midrule
	\textbf{Performance}&&\\
	%\midrule
	&&\\
	Profitable Runs&5&5\\
	Profitable Individuals&50&50\\
	Runs $>$ BH&50&50\\
	Individuals $>$ BH&50&50\\
	Avg Daily Return&$1.08 \times 10^{-3}$&$4.48 \times 10^{-4}$\\
	&\scriptsize{(168\%)}&\scriptsize{(528\%)}\\
	\midrule
	\textbf{Trading Activity}&&\\
	%\midrule
	&&\\
	Number of Trades&2411&1040\\
	&\scriptsize{(32.72\%)}&\scriptsize{(79.87\%)}\\
	%\midrule
	&&\\
	Winning Trades&56.90\%&49.56\%\\
	&\scriptsize{(15.10\%)}&\scriptsize{(40.19\%)}\\
	&&\\	
	%\midrule
	Long Trades&62.69\%&77.26\%\\
	&\scriptsize{(7.74\%)}&\scriptsize{(13.90\%)}\\
	\midrule
	Max Winning&74.46\%&90.77\%\\
	Max Long&74.03\%&89.71\%\\
	\bottomrule
	\end{tabular}
	\caption{Trading activity statistics on the Training set, \textit{Tr} and \textit{TrVa} strategies}
	\label{tab:tset_activity}
\end{table}

\FloatBarrier

\subsection{Out-of-Sample Dataset Performance}
\label{subsec:testresults}
We will now present the results obtained by the same strategies when run on the OoS dataset with the aim of assessing how well they perform on unseen data. The reported quantities and conventions are analogous to those explained in Section \ref{subsec:trainingresults} for the results obtained on the Training dataset.

\begin{figure}[h]
\setlength{\abovecaptionskip}{0pt}
\centering
	\includegraphics[width=0.5\textwidth]{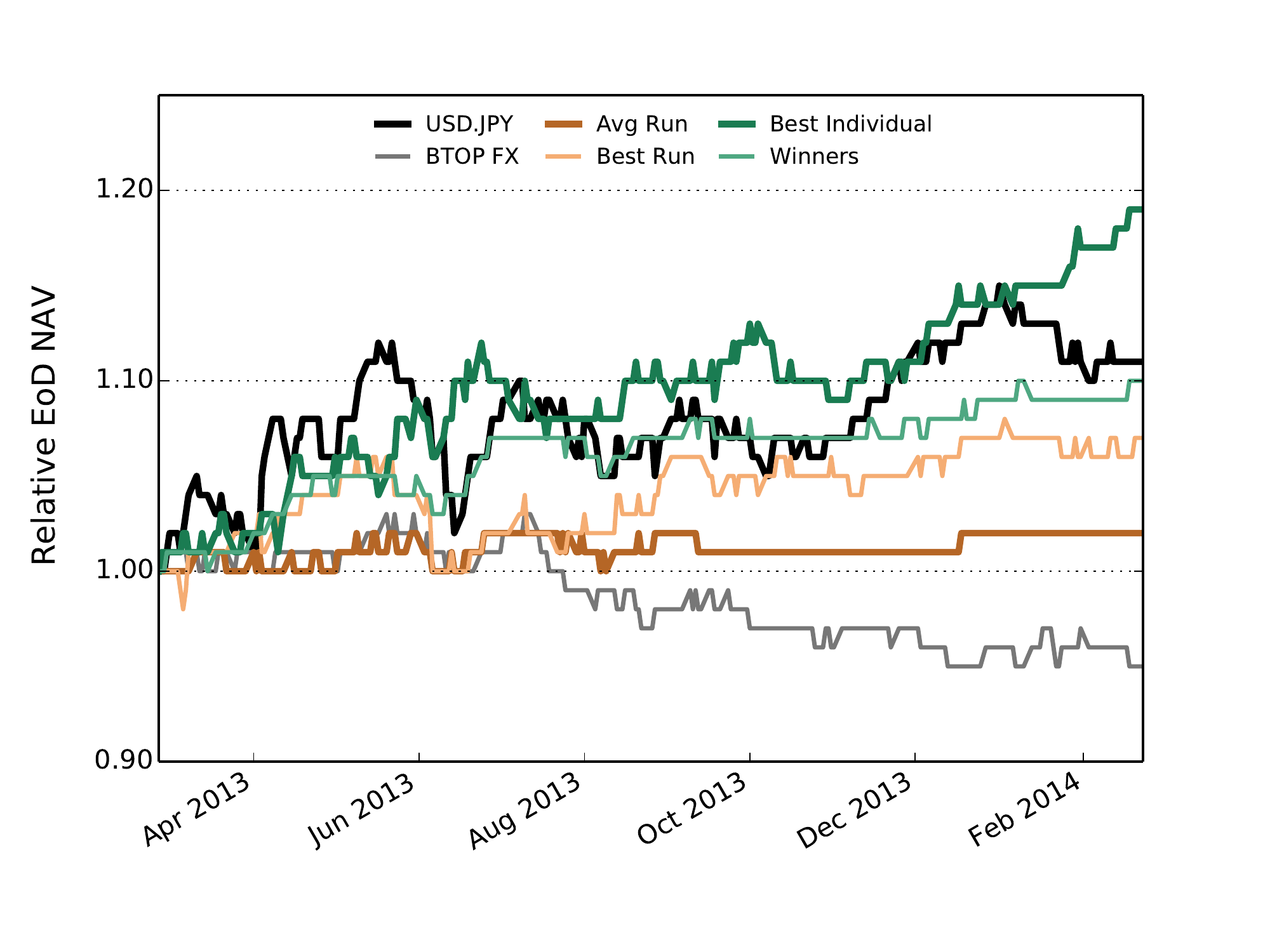}
	\caption{Out-of-Sample set End-of-Day NAV, \textit{Tr} criterion strategies}
	\label{fig:ooset_trainonly}
\end{figure}

\begin{figure}
\setlength{\abovecaptionskip}{0pt}
\centering
	\includegraphics[width=0.5\textwidth]{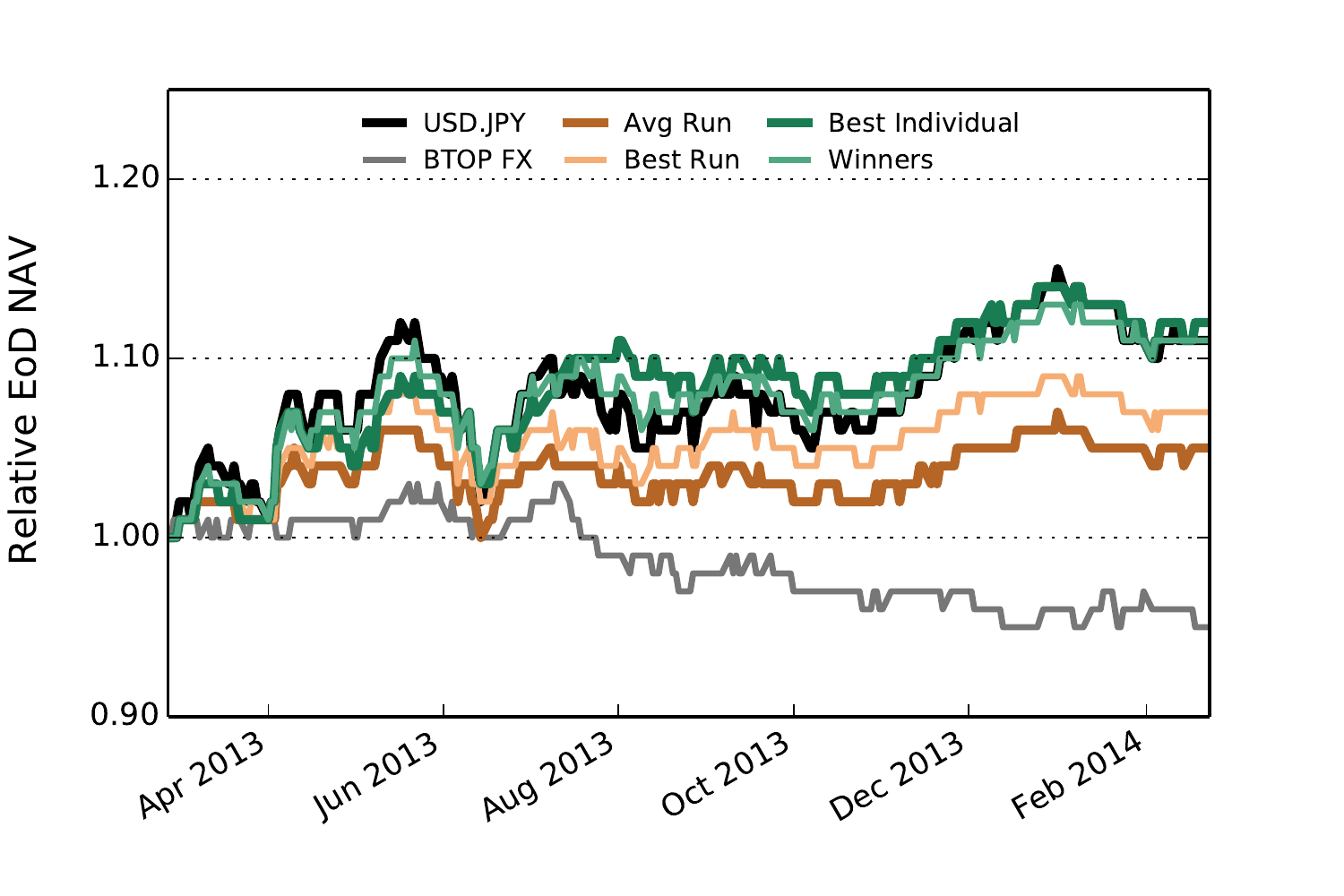}
	\caption{Out-of-Sample set End-of-Day NAV, \textit{TrVa} criterion strategies}
	\label{fig:ooset_traintest}
\end{figure}

Figures \ref{fig:ooset_trainonly} and \ref{fig:ooset_traintest} show the EoD NAV curves, Tables \ref{tab:ooset_trainonly_daily} and \ref{tab:ooset_traintest_daily} feature the daily returns statistics for the two selection criteria and Table \ref{tab:ooset_activity} contains the general performance and trading activity metrics.

\begin{table}
	\centering
	\footnotesize
	\begin{tabular}{lrrr}
	\toprule
		&Return&Days $>0$&$\rho$\\
	\midrule
		\textbf{Benchmarks}&&&\\
	%\midrule
	&&&\\
		USD.JPY&1.11&50.61\%&1.00\\
		BTOP FX&0.95&45.83\%&0.01\\
	\midrule
		\textbf{Runs}&&&\\
	%midrule
	&&&\\
		Run 1&1.07&50.40\%&0.17\\
		Run 2&1.02&51.61\%&0.57\\
		Run 3&0.97&54.44\%&0.13\\
		Run 4&0.96&51.21\%&-0.25\\
		Run 5&1.06&55.65\%&-0.10\\
	\midrule
		%\textbf{Aggregates}&&&&&\\
	%\midrule
		Best Agent&1.19&48.79\%&0.90\\
		Winners&1.10&52.42\%&0.99\\
		Avg Run&1.02&51.82\%&0.98\\
		\bottomrule
	\end{tabular}
	\caption{Daily returns statistics on the Out-of-Sample set, \textit{Tr} strategies}
	\label{tab:ooset_trainonly_daily}
\end{table}

\begin{table}[t]
	\centering
	\footnotesize
	\begin{tabular}{lrrr}
	\toprule
		&Return&Days $>0$&$\rho$\\
	\midrule
		\textbf{Benchmarks}&&&\\
		&&&\\
	%\midrule
		USD.JPY&1.11&50.61\%&1.00\\
		BTOP FX&0.95&45.83\%&0.01\\
	\midrule
		\textbf{Runs}&&&\\
		&&&\\
	%\midrule
		Run 1&1.01&52.02\%&0.93\\
		Run 2&1.05&50.81\%&0.98\\
		Run 3&1.05&50.40\%&0.98\\
		Run 4&1.04&50.40\%&0.97\\
		Run 5&1.04&52.82\%&0.95\\
	\midrule
		%\textbf{Aggregates}&&&&&\\
	%\midrule
		Best Agent&1.12&48.79\%&0.90\\
		Winners&1.11&51.82\%&0.98\\
		Avg Run&1.05&52.42\%&0.99\\
		\bottomrule
	\end{tabular}
	\caption{Daily returns statistics on the Out-of-Sample set, \textit{TrVa} strategies}
	\label{tab:ooset_traintest_daily}
\end{table}

\begin{table}[h]
	\centering
	\begin{tabular}{lll}
	\toprule
	&\textit{Tr}&\textit{TrVa}\\
	\midrule
	\textbf{Performance}&&\\
	%\midrule
	&&\\
	Profitable Runs&3&5\\
	Profitable Individuals&26&42\\
	Runs $>$ BH&0&0\\
	Individuals $>$ BH&4&1\\
	Avg Daily Return&$7.67 \times 10^{-5}$&$1.89 \times 10^{-4}$\\
	&\scriptsize{(2538\%)}&\scriptsize{(2128\%)}\\
	\midrule
	\textbf{Trading Activity}&&\\
	%%\midrule
	&&\\
	Number of Trades&3247&1460\\
	&\scriptsize{(29.57\%)}&\scriptsize{(96.56\%)}\\
	%%\midrule
	&&\\
	Winning Trades&56.97\%&48.95\%\\
	&\scriptsize{(21.33\%)}&\scriptsize{(38.43\%)}\\
	&&\\
	Long Trades&57.49\%&76.64\%\\
	&\scriptsize{(17.70\%)}&\scriptsize{(17.23\%)}\\
	\midrule
	Max Winning&83.45\%&86.70\%\\
	Max Long&75.13\%&90.99\%\\
	\bottomrule
	\end{tabular}
	\caption{Trading activity statistics on the Out-of-Sample set, \textit{Tr} and \textit{TrVa} strategies}
	\label{tab:ooset_activity}
\end{table}
%\FloatBarrier

\subsection{Tree Structure}
\label{subsec:treestructure}

To investigate the difference in the logic of the trading strategies selected under the two criteria, we analyzed the actual symbolic expression trees encoding the individuals.
In Table \ref{tab:treelength} we report the average tree length and number of variable nodes computed across the total 50 strategies per criterion.

\begin{figure}
\setlength{\abovecaptionskip}{0pt}
\centering
\includegraphics[width=0.5\textwidth]{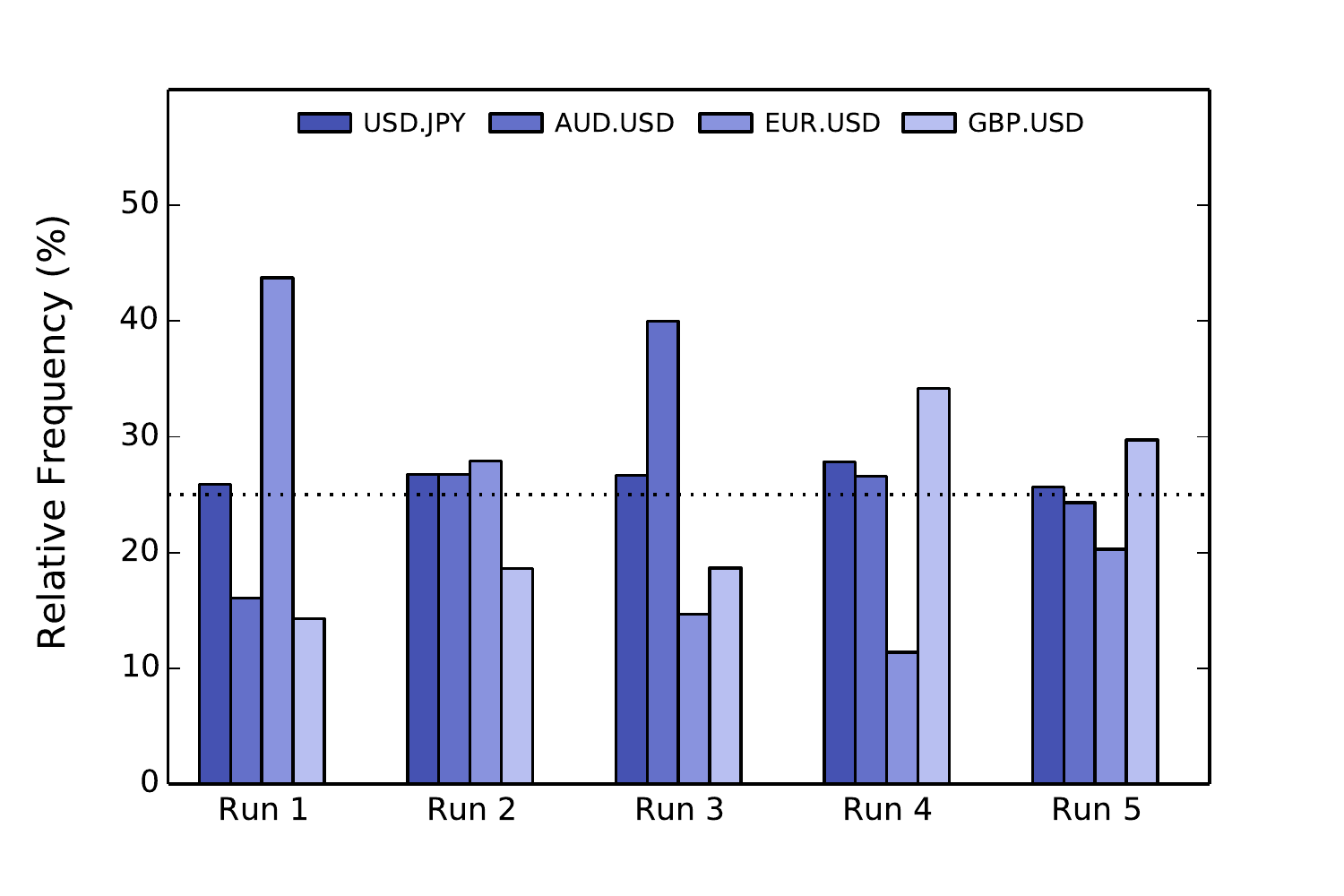}
\caption{Trees for \textit{Tr} strategies, currencies' relative frequencies}
\label{fig:trainonly_currency}
\end{figure}

\begin{figure}
\setlength{\abovecaptionskip}{0pt}
\centering
\includegraphics[width=0.5\textwidth]{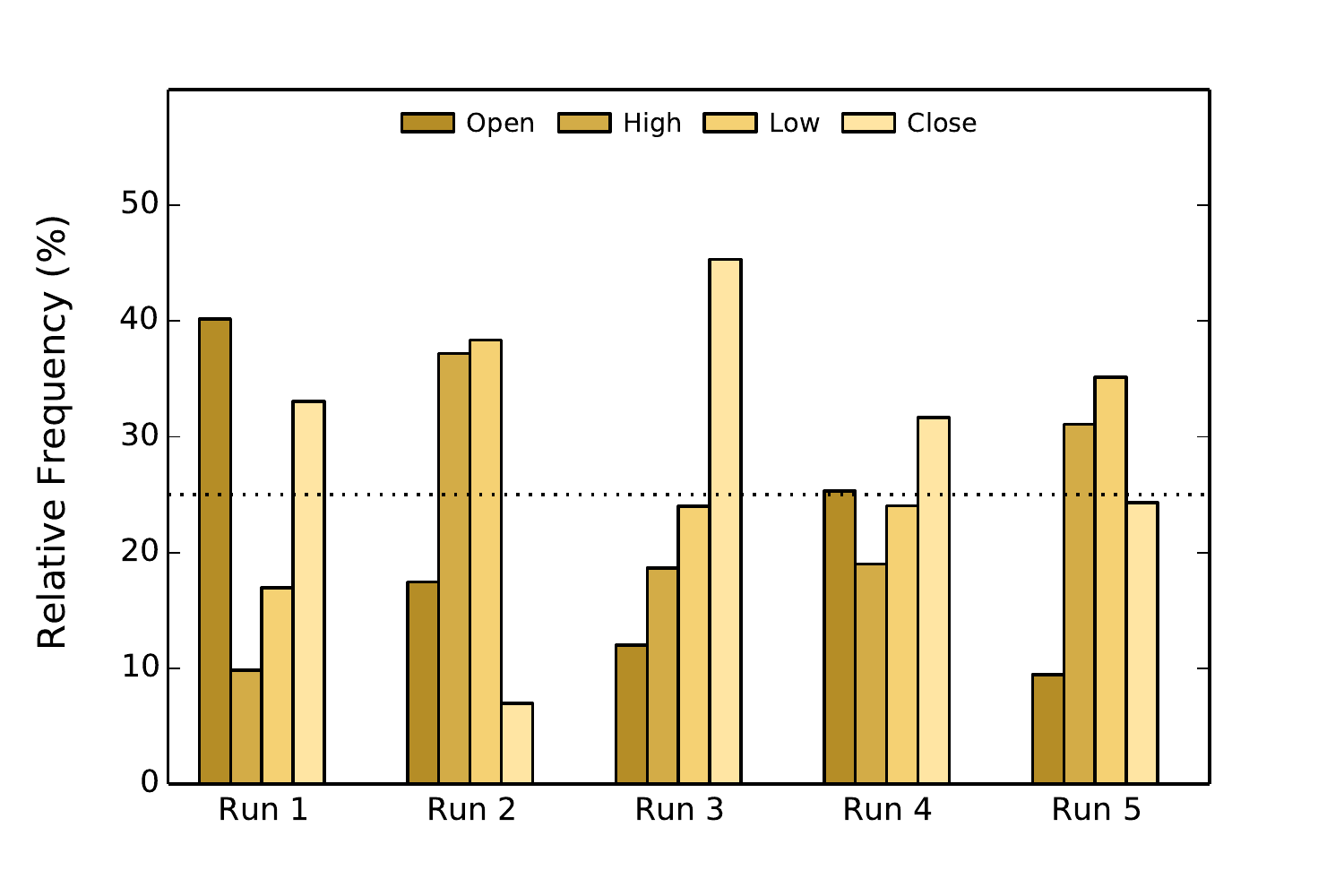}
\caption{Trees for \textit{Tr} strategies, OHLC values' relative frequencies}
\label{fig:trainonly_ohlc}
\end{figure}

Another interesting aspect to analyze is the relative frequencies of the input variables within the expression trees. Figures \ref{fig:trainonly_currency} and \ref{fig:trainonly_ohlc} show, respectively, the per-run relative frequencies of the currency and bar value (Open, High, Low, Close) of the variable nodes for the trees selected under the \textit{Tr} criterion. Figures 8 and 9 instead display them for strategies from the \textit{TrVa} criterion.

\begin{table}
	\centering
	\footnotesize
	\begin{tabular}{lll}
	\toprule
	&\textit{Tr}&\textit{TrVa}\\
	\midrule
	Tree Length&34.06&31.68\\
	&\scriptsize{(2.96)}&\scriptsize{(3.97)}\\
	&&\\
	Symbols Count&8.25&7.12\\
	&\scriptsize{(1.5)}&\scriptsize{(0.8)}\\
	\bottomrule
	\end{tabular}
	\caption{Average and standard deviation values of the tree length and variable symbols count, \textit{Tr} and \textit{TrVa} strategies}
	\label{tab:treelength}
\end{table}

\begin{figure}[t]
\setlength{\abovecaptionskip}{0pt}
\centering
\includegraphics[width=0.5\textwidth]{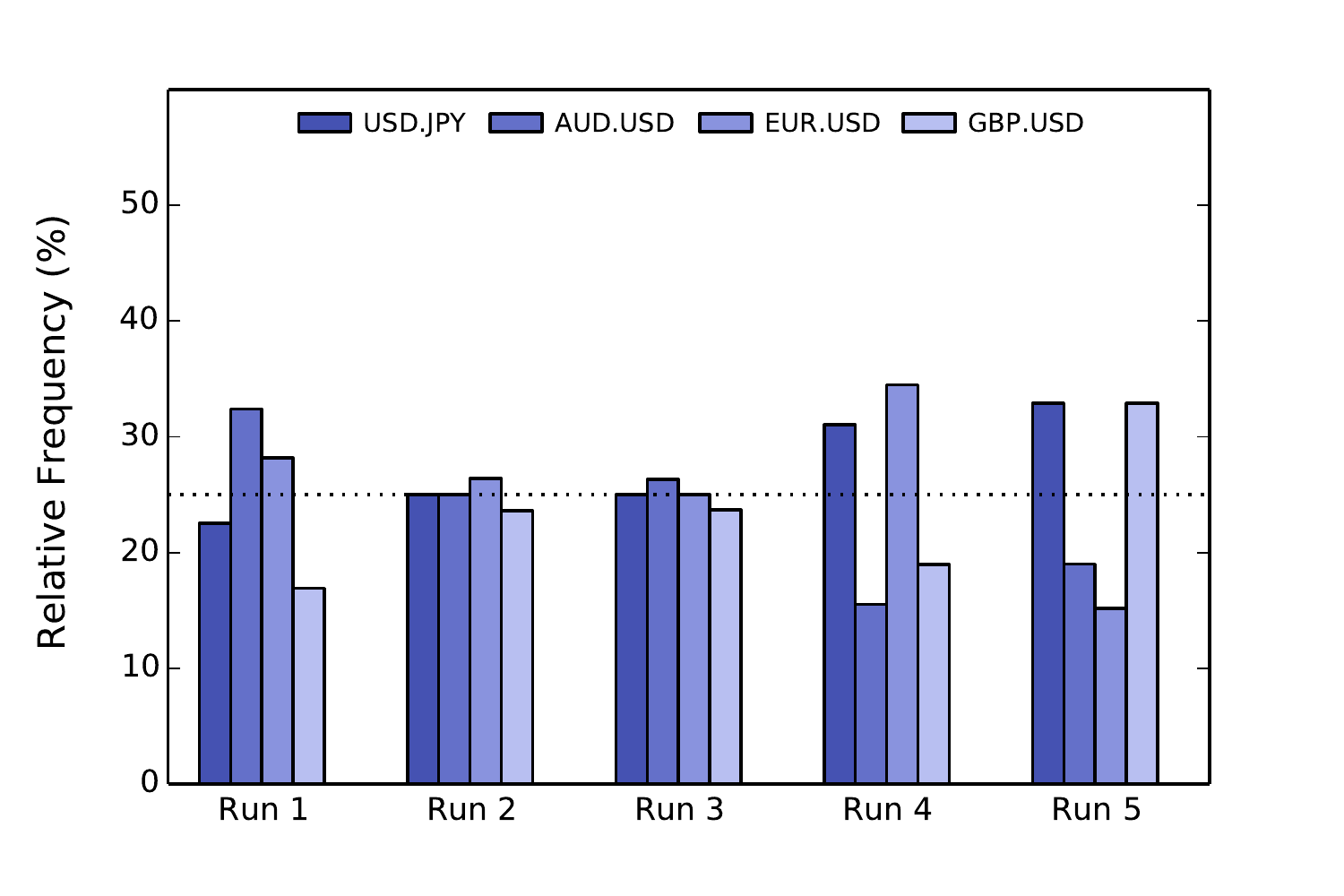}
\caption{Trees for \textit{TrVa} strategies, currencies' relative frequencies}
\label{fig:traintest_currency}
\end{figure}

\begin{figure}[t]
\setlength{\abovecaptionskip}{0pt}
\centering
\includegraphics[width=0.5\textwidth]{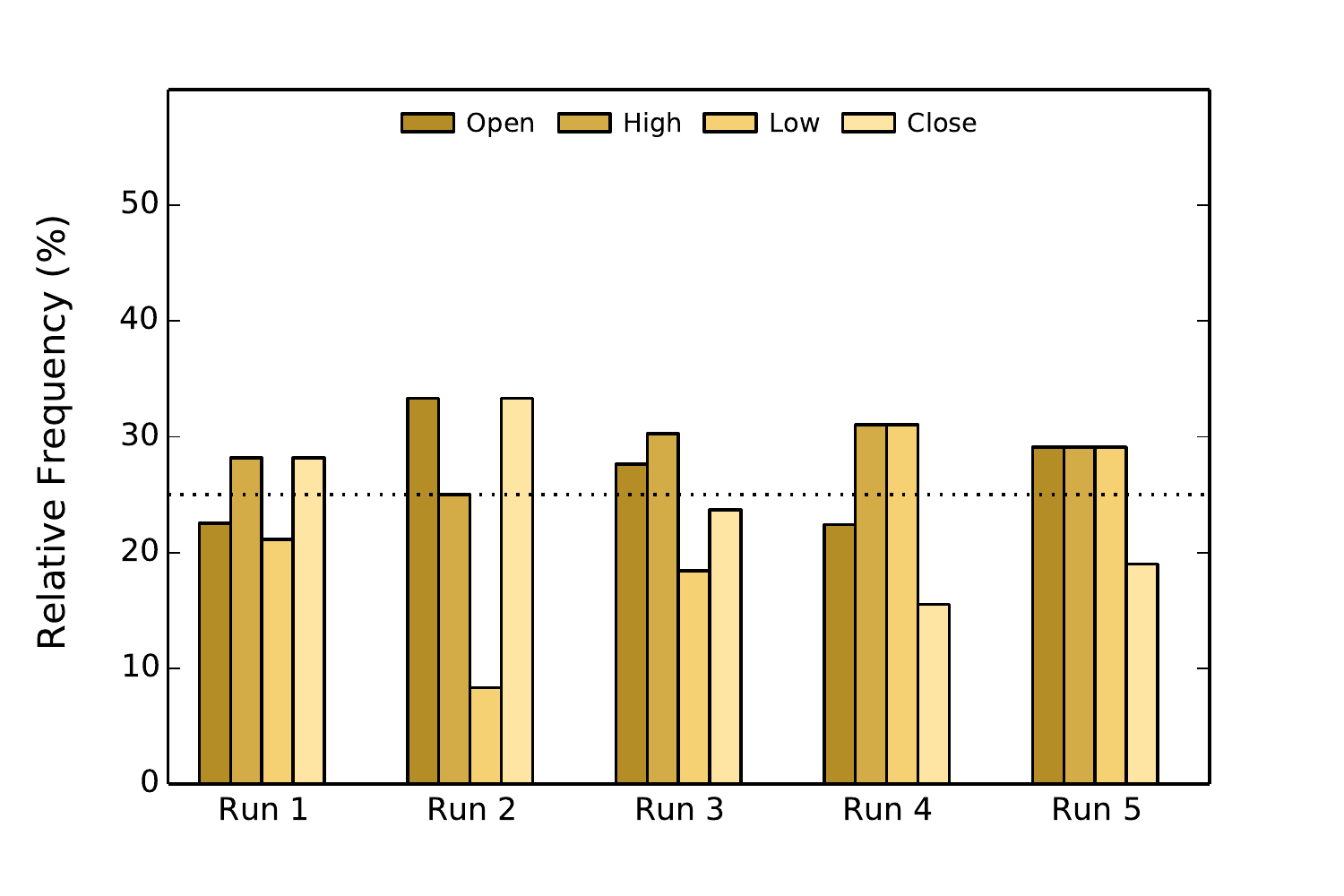}
\caption{Trees for \textit{TrVa} strategies, OHLC values' relative frequencies}
\label{fig:traintest_ohlc}
\end{figure}

%\FloatBarrier

%% file: 5_Discussion.tex
\section{Discussion}
\label{sec:discussion}
In this Section we will first analyze the strategies performance on the Training and Out-of-Sample datasets. Then we will characterize the common aspects across the two datasets of the strategies selected with the \textit{Tr} and \textit{TrVa} criteria, with a focus on the Out-of-Sample set. Successively we will examine differences in strategy structure and variable selection occurring under the two criteria. Lastly we will perform a results comparison with similar previous investigations.

\subsection{Training Dataset}
\label{subsec:trainingset_discussion}
Performance on the Training dataset is overwhelmingly positive, with all of the strategies for both selection criteria achieving profitability and beating the two benchmarks: a return of 5\% for USD.JPY and 1\% for BTOP FX.
The best performing agent comes, unsurprisingly, from the Training only criterion, obtaining a final return of 32\%. The average return is instead 26\% for \textit{Tr} strategies and 16\% for \textit{TrVa} ones.

Examining the daily return series, both USD.JPY and BTOP FX benchmarks exhibit behavior very close to a random walk,  with 51.20\% and 51.50\% of profitable daily returns respectively. Evolved strategies, especially from \textit{Tr}, instead show a solid day-to-day profit potential with a single strategy maximum of 66.20\% ($p=1.31 \times 10^{-6}$ under the null hypothesis the daily return series follows a 50/50 positive/negative distribution); for \textit{TrVa} this value is lower, 56.81\% ($p=0.027$) but still very improbable under the null hypothesis.

A very interesting aspect emerging for both \textit{Tr} and \textit{TrVa} is that the best daily profitability ratio is not given by individual strategies, but by aggregates: the global average for \textit{Tr}, 75.12\%, and the average of the Run 1 strategies for \textit{TrVa}, 60.09\%; the average of the 5 runs' top performers also outperforms the best single strategy for both criteria.
This suggests that the evolved strategies present, to some extent, complementary trading logic. If day-to-day losses are to be minimized then, suitable strategies could be combined to form the basis for a trading portfolio.

Considering trading activity \textit{Tr} strategies are on average 2.3 times more active than the \textit{TrVa} ones, but both have high, active, trading profiles: 2411 total trades for \textit{Tr} or, considering the Training period length of 213 days, 11.3 trades per day and 1040 trades, or 4.8 trades per day, for \textit{TrVa} strategies.

Analyzing the trade accuracy, the ratio of profitable trades performed, we find that \textit{Tr} achieves a value of 56.90\% and \textit{TrVa} of only 49.56\%. However, the standard deviation of the accuracy values of \textit{TrVa} strategies is much higher and surprisingly the individual strategy with the highest trade accuracy, a staggering 90.77\% ($p=4.98 \times 10^{-205}$) was selected under this criterion while the most accurate \textit{Tr} strategy "only" has an accuracy of 74.46\% ($p=1.02 \times 10^{-162}$).

Regarding market side exposure, both criteria selected strategies presenting a strong bias towards the Long side, however the reasons for this can be explained by the dynamics of the traded instrument price series which also present a Long side trend.

\subsection{Out-of-Sample Dataset}
\label{subsec:ooset_discussion}
On the Out-of-Sample dataset performance is still generally solid, with more than half of the strategies and run averages obtaining profitable returns. Understandably, in this case the obtained returns are lower than on the Training dataset, with the best strategy achieving a 19\% return and averages obtaining 2\% and 5\% for \textit{Tr} and \textit{TrVa} respectively.

On this dataset however, while the best performer was still found among the strategies selected under \textit{Tr}, the total number of agents achieving profits as well as the averaged results are better for the \textit{TrVa} criterion.
All of the profitable strategies for both criteria beat the BTOP FX benchmark while only 4 strategies for \textit{Tr} and 1 for \textit{TrVa} manage to beat the buy-and-hold USD.JPY approach.

Here the daily return profitability is lower than on the Training set as the average values are generally in the neighborhood of 50\% for both criteria. Once again the better performers for this metric are strategy aggregates: 54.44\% ($p=0.094$) for the \textit{Tr} Run 3 average, 52.82\% ($p=0.20$) for the \textit{TrVa} Run 5 average.

On the OoS dataset \textit{Tr} strategies are 2.2 times more active than \textit{TrVa} ones. Trades per day are 12.8 for \textit{Tr} and 5.7 for \textit{TrVa}.

Trade accuracy average is 56.97\% for \textit{Tr} and 48.95\% for \textit{TrVa}. On this dataset as well the best trade accuracy, 86.70\%, ($p=2.6 \times 10^{-26}$) comes from a \textit{TrVa} strategy.

The market side exposure again presents a preference for the Long side.

Examining trade activity, accuracy, and side exposure we find that these results are very similar to the ones obtained on the Training set for both criteria. This indicates that while the obtained profits on unseen data are lower, other important strategy performance indicators are preserved, providing the financial practitioner with a consistent basis for which strategies to choose for live trading.

\begin{figure}
\setlength{\abovecaptionskip}{0pt}
\centering
	\includegraphics[width=0.5\textwidth]{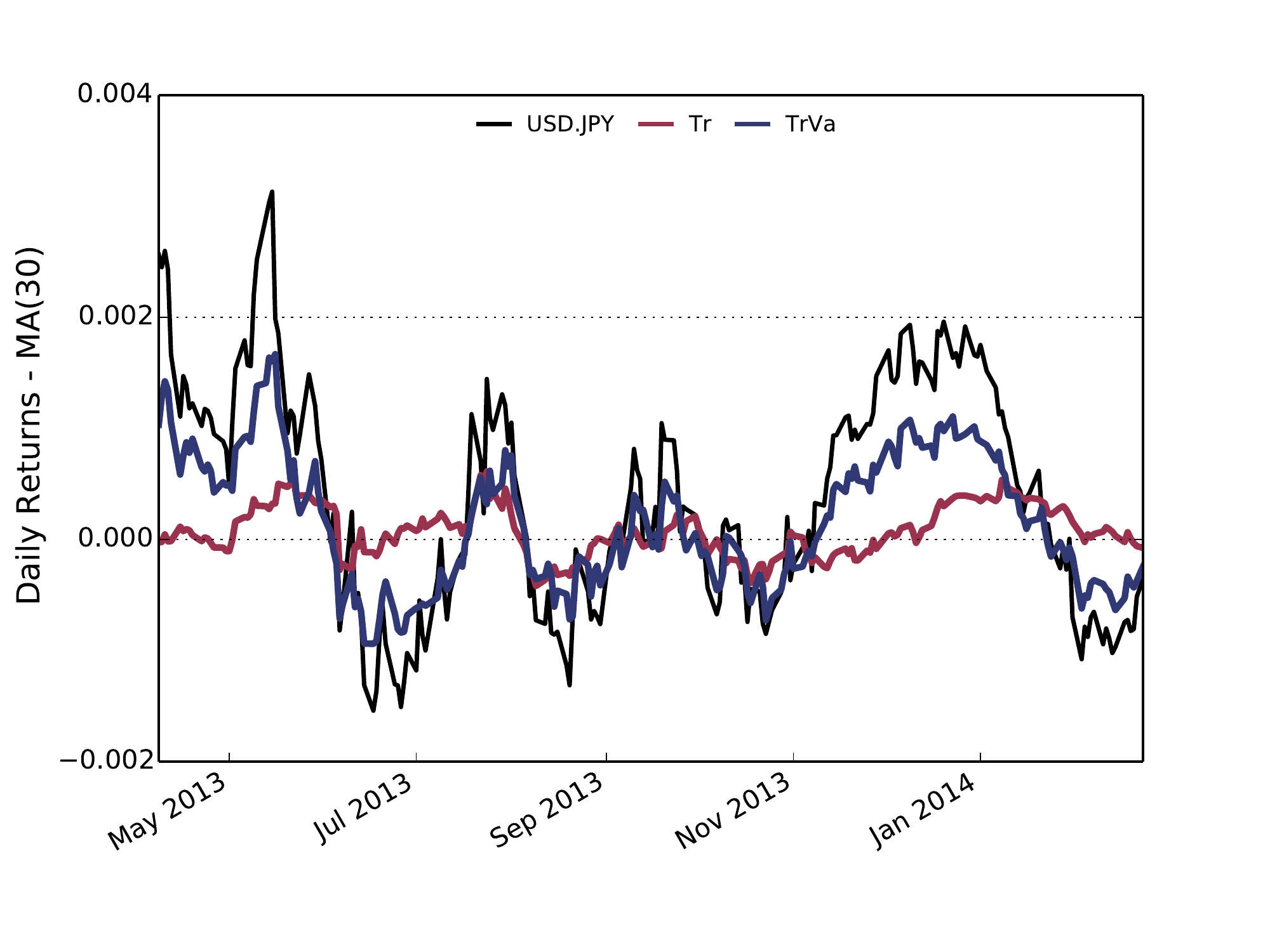}
	\caption{30-Days moving averages for the Out-of-Sample daily returns series of the USD.JPY, \textit{Tr} average, and \textit{TrVa} average}
	\label{fig:mareturns}
\end{figure}

Figure \ref{fig:mareturns} shows the 30-days moving averages for the daily returns of USD.JPY and the 50 strategies' averages for \textit{Tr} and \textit{TrVa}. We note that for both of the criteria this quantity, despite exhibiting oscillations correlated with the underlying instrument, does not show stable decreasing trends during the course of the Out-of-Sample dataset. This indicates that our overall approach based multi instrument input data without using established technical indicators for model building succeeds at learning patterns which maintain performance levels for up to one year after being generated without any sort of model retraining in place. Another conclusion is that the usage of unprocessed price quotes, at least for the total considered timespan of two years, also does not lead to performance degradation and that our GP setup is able to cope with the nonstationarity of its input data.

\subsection{Selection Criteria\\Performance Differences}
\label{subsec:criteria_discussion}
Analyzing the strategies selected by the two criteria and their performance on Training and Out-of-Sample datasets it emerges that the two criteria select strategies exhibiting different behaviors and that such characteristics are preserved when run on unseen data.

From the point of view of pure profits, the highest yielding strategies were always found via \textit{Tr}, \textit{TrVa} instead resulted in better averaged results.

Pearson correlation coefficients with the return series of the USD.JPY instrument prices together with the average proportion of long-side trades as well as a simple qualitative inspections of the charts in Figure \ref{fig:tset_trainonly}, \ref{fig:tset_traintest}, \ref{fig:ooset_trainonly}, and \ref{fig:ooset_traintest} show that the \textit{TrVa} criterion consistently selects agents much more correlated to the security they are trading.

Comparing the activity statistics, \textit{Tr} reliably selects strategies trading more often while the \textit{TrVa} criterion results in strategies on average performing less trades but individually spanning a larger range of activity levels.

Regarding trade accuracy, on average \textit{Tr} produces better strategies but under \textit{TrVa} is where the individual strategies having the highest accuracy were found.

On the Training Dataset, it's \textit{Tr} strategies having better average daily returns while on the Out-of-Sample data not only this value for \textit{TrVa} is more than double but also its standard deviation is slightly lower. Moreover, on OoS it's \textit{TrVa} obtaining a higher day-to-day profitability ratio. This fact, together with the higher number of non-profitable strategies as well as the higher average trade accuracy exhibited under \textit{Tr}, leads us to conclude that when \textit{Tr} strategies lose on the Out-of-Sample dataset, they lose more than their counterparts selected with \textit{TrVa}.

Drawing conclusions on the performance of strategies from the two criteria on the Out-Of-Sample dataset, both succeeded in evolving strategies successfully maintaining the trading activity and accuracy levels shown on the training data. Considering that our signal-based approach requires strategies to alter their output value to trigger an execution, this means the agents successfully generalized patterns present in the input price series to a good level while not fixating on features specific to the training dataset with no or very little applicability on other data even when bred exclusively on the training set. Therefore we conclude our approach exhibits very low overtraining.

From the point of view of profitability, both criteria successfully produce strategies profitable on unseen data, however chances under \textit{TrVa} are better. 
Attempting to define the general characteristics of the strategies selected from the two criteria we can conclude that the \textit{Tr} criterion, statistically, results in more volatile individuals, with higher trading activity and profit potential but also with an increased chance of substantial losses. The \textit{TrVa} criterion instead results in more modest, but far less risky, profits with minimal chance of loss and with very high correlation to the underlying traded instrument.

Therefore both criteria are, in our own opinion, suitable for the generation of strategies to be applied on actual, live markets. The amount of investment risk that can be sustained or desired would favor the application of one over the other.
Moving from the single strategy level to the portfolio level, we have also seen that both criteria produce strategies with complementary logic that if successfully combined would lower risk even more.

Speculating on the nature of patterns individuated and learned by the two criteria, we can posit that strategies selected by the \textit{Tr} criterion are able to better individuate and exploit profitable long-term trends present in the data. By being stable, these trends allow for more frequent trading and over time they exhibit little correlation with the underlying instrument. On the other hand, \textit{TrVa} would seem to favor patterns on a much shorter timescale, over time leading to lower risk and maintaining a very high correlation with the instrument.

\subsection{Strategy Structure and Variable Selection}
\label{subsec:varselection_discussion}
From the values reported in Table \ref{tab:treelength} we conclude that there is no significant difference in total tree length or number of featured variable symbols in the strategies selected by the two criteria.

Analyzing the relative frequencies with which the different input currencies appear in the selected strategies, displayed in Figures \ref{fig:trainonly_currency} and \ref{fig:traintest_currency}, we notice that while the global average frequencies for all four currencies are very close to the expected value of 25\%, the individual runs present different variable usage "spectra".

The traded instrument, USD.JPY, is the only one that for every \textit{Tr} and \textit{TrVa} run deviates the least from the expected value, exhibiting a relative standard deviation of 3\% for \textit{Tr} runs and of 16\% for \textit{TrVa} runs. This indicates that our approach successfully and reliably recognizes the importance of the traded instrument even with a fairly high-dimensional input space as the one we make use of.

Frequencies for other instruments instead vary more between runs, with relative standard deviations of 32-57\% for \textit{Tr} runs and 16-28\% for \textit{TrVa}.
For some runs the selected strategies make more or less and equal use of all of them, in others the evolutionary process converges towards a pool focused on only one or two other instruments besides the traded one.
This could be interpreted as the separate runs finding patterns mainly involving a subset of the security basket while relatively disregarding other inputs.

The always lower relative standard deviations of \textit{TrVa} versus \textit{Tr} indicate that the former criterion, due to its implicit focus towards finding more general patterns, is less prone to search space exploitation, or breeding of super-specialists, and therefore maintains a more average variable usage distribution across runs. This is consistent with the better generalization capabilities on unseen data shown by \textit{TrVa} from a profitability perspective, as detailed in Section \ref{subsec:criteria_discussion}. 

Regarding the distribution of the selected variables with respect to the Open, High, Low, and Close values, there appear to be no significant patterns or differences among the two criteria.

%\subsection{Investment Considerations}
%\label{subsec:investment_discussion}

\subsection{Comparison with Previous Work}
\label{sec:prevwork_discussion}
Performing a comparison between works in this field is not straightforward, given the many variable aspects involved in automated trading strategies and systems. We will then here examine only the previous works in the literature with a similar setup to our own: utilizing GP or GA to evolve strategies or models, trading on foreign exchange, utilizing unlevered returns, and including transaction costs in the evaluation process.

The next-day return forecasting models of \cite{vasilakis2012genetic} obtains, after transaction costs, an annualized return of 5.9\% on the EUR.USD. Despite our strategies having a higher trading activity, and being therefore more exposed to transaction costs, their best result is close to our average Out-Of-Sample result of 5\% but considerably lower than our best one, 19\%. 

The hybrid GA - reinforcement learning technique proposed in \cite{hryshko2006development} utilizes 5 minutes data as we do. They report profits of 6\% on the EUR.USD in 3.5 months of out-of-sample data, which are extrapolated to 20\% annualized returns. That is slightly better than our overall best performer, however considering that it has been shown that returns tend to degrade over time \cite{dempster2001real} it is unknown how well their generalized figure would hold in practice.

The work described in \cite{wilson2010interday2} is the one utilizing the setup the most similar to our own as they are the only other ones evolving free-form agents; also similarly, their work and ours show a very high best trade accuracy: 90\% or higher. Regarding profits obtained on the USD.JPY, their raw fitness achieves 4.27\% and their conservative one 13.44\%. Their conservative fitness is based on the Stirling ratio and it aims at explicitly reducing the strategy losses while their raw fitness is the unmodified value of assets held and is the same as our fitness. Therefore it can be said that our results are better than theirs as our non risk-mitigating fitness on average performs slightly higher then theirs, but ours in the best case obtains better returns than their risk-mitigating one. 
However, final profits are strongly tied to the chosen evaluation period so they are not fully comparable between different works. Even disregarding the actual obtained profit figures, we produced very similar results without data preprocessing or rolling window retraining; techniques which reduce the complexity of the input space and allow model retraining.

Another investigation with aspects similar to our own is the one from Manahov and Hudson \cite{manahov2013new}, as they employed 5 minutes data as well as a very high population approach. Their average trade accuracy on out-of-sample data is close to the one we obtained, a few points in excess of 50\%. However, we are not able to compare the obtained results because they express and report their profits as excess returns against a US Treasury Bill and their out-of-sample period is only 1 month long.

Even comparing with other, more dissimilar approaches, our results appear better, more resilient, or higher: in contrast to \cite{neely2003intraday} and \cite{mendes2012forex} we do not struggle at all in the presence of transaction costs while \cite{dempster2001real} reports an annualized return of 7\% on the GBP.USD but their performance substantially degrades after nine months into the out-of-sample period. 

In general, we are able to state that against previous related work we successfully managed to improve on several weaknesses. Our approach, while still showing lower profits on unseen data as opposed to its training set, maintains essentially analogous trade frequencies and accuracies between the two datasets, indicating successful and generalizable learning of profitable patterns. Additionally, by factoring in transaction costs already from the breeding process our evolved strategies are fully and stably profitable, both in and out of sample, in their presence despite their high exhibited trading activity: on average 4 to 13 trades per day. Trade accuracy ranges from statistically expected values to very high ones, pointing to the conclusion that explicitly including risk mitigation/loss aversion measures in the fitness function, something we do not have, is not strictly necessary to evolve strategies with very little losses.
Lastly, despite not employing any sort of periodic or performance-triggered model retraining and making use of raw price quotes, we were able to successfully generate models which do not exhibit signs of diminishing returns on out-of-sample performance for up to one year. This contrasts with what \cite{hryshko2006development} reports and establishes that neither the use of common technical indicators nor data preprocessing are required for the emergence of successful, profitable strategies. 

From the point of view of out-of-sample profits, the results of our approach outperform most documented attempts and in the worst case they are on-par with the best results found in literature.

%% file: 6_Conclusions.tex
\section{Conclusions}
\label{sec:conclusions}
We described, implemented and tested a genetic programming system for the evolution of currency trading strategies in the foreign exchange market. The proposed system introduced several innovative aspects aimed at facilitating the application of the strategies to live, production environments as well as finding trading patterns breaking away from traditional technical analysis models.

The principal innovation is constituted by having price data from multiple currency pairs in addition to the single one being traded as inputs to the system. This successfully resulted in the evolution of strategies featuring multi-currency trading logic.

We examined performance of two different sets of solutions, obtained across several independent runs, over one year of out-of sample, unseen data: one set selected according to performance on a single training dataset, and the second one making use of a combined performance metric involving an additional validation set. For all the performed runs profitable strategies were found, indicating the reliability of our proposed architecture.

The two selection criteria both resulted in best performing individuals maintaining an active trading profile (6 to 13 trades per day), having a remarkable trade accuracy (83 to 87\% of performed trades are profitable), as well as achieving final profits competitive with prior documented investigations (12 to 19\%). Both criteria appear therefore suitable for live application.
However, the strategies selected under the two criteria also exhibit some differences: the training only selection policy appears to be more suited if looking for strategies maximizing profits and, therefore, risk and having a NAV profile lowly correlated with the price of the underlying instrument. The combined selection criterion instead is better at finding agents minimizing risks at the expense of the profitability potential as well as having high correlation with the underlying security.

The system also appears able, over multiple runs, to evolve strategies exhibiting complementary trading logic which can be combined to minimize risk even further without impacting the profit potential.

\subsection{Future Directions}
\label{subsec:future}
The proposed system was tested on intraday foreign exchange data with a 5 minutes frequency, however our technique and setup can be utilized to trade other asset classes (stocks, ETFs, \ldots), or on other datapoint frequencies without further modification.

The fitness metric utilized in this work is based on the obtained final return, which is dependant on dataset length. The combined selection criterion ties together the fitness scores obtained on the training and validation datasets. As in the presented evaluations the training and test sets had different lengths, despite still showing promising results, the combined criterion's efficacy at learning less risky patterns traded-off with its maximum profit potential. In order to better assess its validity it would be therefore necessary to evaluate it in a scenario where training and test sets have the same length. Alternatively, this issue could be addressed by defining a different, dataset length invariant, fitness score.

Considering the variety of trading models utilizing autoregressive components in both literature and industry practice, it is worth investigating the performance impact resulting from the introduction of past security prices to the pool of variables the Genetic Programming strategies have access to. 